\documentclass{article} 
\usepackage{iclr2021_conference,times}

\usepackage{amsmath,amsfonts,bm}

\def\eqref#1{equation~\ref{#1}}

\def\1{\bm{1}}

\DeclareMathAlphabet{\mathsfit}{\encodingdefault}{\sfdefault}{m}{sl}
\SetMathAlphabet{\mathsfit}{bold}{\encodingdefault}{\sfdefault}{bx}{n}

\usepackage{hyperref}
\usepackage{url}

\usepackage{ctable}
\usepackage{subcaption}

\title{Transitioning from Real to Synthetic data:\\ Quantifying the bias in model}

\author{Aman Gupta, Deepak Bhatt \& Anubha Pandey\\
Mastercard, India \\
\texttt{\{aman.gupta,deepak.bhatt,anubha.pandey\}@mastercard.com} 
}
\iclrfinalcopy 
\begin{document}

\maketitle

\begin{abstract}
With advent of generative modelling techniques, synthetic data and its use has penetrated across various domains from unstructured data such as image, text to structured dataset modelling healthcare outcome, risk decisioning in financial domain and many more. It overcomes various challenges such as limited training data, class imbalance, restricted access to dataset owing to privacy issues. To ensure trained model used for automated decisioning purposes makes fair decision there exist prior work to quantify and mitigate those issues. This study aims to establish trade-off between bias and fairness in the models trained using synthetic data. Variants of synthetic data generation techniques were studied to understand bias amplification including differentially private generation schemes. Through experiments on a tabular dataset, we demonstrate there exist varying level of bias impact on models trained using synthetic data. Techniques generating less correlated feature performs well as evident through fairness metrics with 94\%, 82\%, and 88\% relative drop in DPD (demographic parity difference), EoD (equality of odds) and EoP (equality of opportunity) respectively, and 24\% relative improvement in DRP (demographic parity ratio) with respect to the real dataset. We believe the outcome of our research study will help data science practitioners understand the bias in use of synthetic data.
\end{abstract}

\section{Introduction}

Machine learning systems are being adopted in every aspect of our daily life. Amazon and Netflix uses it for movie recommendations, recommending products customized based on user preference. Among many, recruiting industry is one such example leveraging AI to find potential candidates. Owing to its capability to reduce manual hours and assist in automated decision-making, they are being widely deployed and integrated across various use-cases such as public use, policy-making, predictive policing, etc. On the other hand, though machine learning systems are quite successful yet it found that ``Robots are racist and sexist, just like the people who created it\footnote{https://www.theguardian.com/commentisfree/2017/apr/20/robots-racist-sexist-people-machines-ai-language}", claimed that ``crime rate in color people are higher\footnote{https://www.propublica.org/article/machine-bias-risk-assessments-in-criminal-sentencing}" and ``dark skins are unattractive\footnote{https://www.theguardian.com/technology/2016/sep/08/beauty-contest-doesnt-like-black-people}", ``recommending less qualified male candidates higher than more qualified female candidates on job portal" ~\citep{lahoti2019ifair}. As machine learning models learn what humans teach them, therefore, like people, machine learning systems are vulnerable to amplify existing human biases and societal stereotypes when they make decisions~\citep{wang2019balanced}. One of the factors that may influence the fairness of machine learning models is change in model outcome with protected attributes like gender, race, religion, sexual orientation, economic condition, etc~\citep{mehrabi2019survey}. Recent research found that even if we don't use the protected attributes for the model training, information about the race, gender, and age are implicitly encoded into intermediate representations of the model~\citep{elazar2018adversarial}. 

Various bias mitigation techniques exists such as: (1) Pre-processing: transforming data representation before training~\citep{d2017conscientious}~\citep{bellamy2018ai}, (2) In-processing: modify the existing algorithms during the training process to remove discrimination, and (3) Post-processing: transform the model output to improve prediction fairness  ~\citep{berk2017convex}. The effectiveness of these methods have been tested on a real dataset; however, synthetic data is gaining traction in light of limited data access arising due to privacy and compliance. Recent research aims to study how biases are amplified, for example, in compressed deep neural networks~\citep{hooker2020characterising}. Another study estimates the impact of bias in adopting differentially private mechanisms to protect the model from learning sensitive attributes~\citep{bagdasaryan2019differential}.

In this study, we are particularly interested in bias amplification when models are trained using synthetic data. Generative adversarial networks (GANs) have become a popular choice for synthetic data generation~\citep{goodfellow2014generative}. Several variants of GANs are available today that are more stable and capable of generating realistic samples. In this paper, we have used state-of-the-art GAN architechture for tabular data generation: CTGAN~\citep{xu2019modeling}, Gaussian Copula GAN ~\citep{masarotto2012gaussian}, and Copula GAN\footnote{https://en.wikipedia.org/wiki/Copula\_\%28probability\_theory\%29}.  Further to generate differentially private synthetic data, PATE-GAN~\citep{jordon2018pate} is used. 

The major contributions are: (1) demonstrating the impact of bias on models trained using synthetic data (2) Studied how differentially private synthetic data affect the model accuracy and the fairness (3) compared and contrast the key differences in model trained using synthetic data and its differentially private version.

\section{Experimental Results}

\subsection{Fairness Metrices}
There exists a myriad of notions in the literature to quantify fairness. Each measure emphasizes different aspects of fairness. Fairness in machine learning measures the degree of disparate treatment for different groups (e.g., female vs. male), or individual fairness, emphasizing similar individuals should be treated similarly. Evaluation metrics to measure group fairness are (i) Demographic Parity Difference (DPD)~\citep{verma2018fairness}. (ii) Demographic Parity Ratio (DPR) (iii) Equality of Odds (EoD) (iv) Equality of Opportunity (EoP)~\citep{hardt2016equality}. Please find the details of the metrics in the supplementary. We have used FairLearn library\footnote{https://fairlearn.github.io/} to access fairness metrics. Some cut-off values have been defined in the literature for these metrics, suggesting whether models are fair or not. For DPD, EoD, and EoP, if the absolute value is smaller than 0.1, the model can be considered fair, whereas for DPR fairness range for this metric is between 0.8 and 1.25.

\subsection{Dataset Description}
We have used Adult Dataset\footnote{http://archive.ics.uci.edu/ml} for our analysis, a well-known example, which is widely used in the fairness literature to predict whether the income of an individual exceeds $ \$50K/yr$ based on census data. The Adult dataset has about 48K records with a binary label indicating a salary $\leq\$50K$ or $\geq\$50K$, with a class imbalance of $[76\%, 24\%]$, where 76\% of the observations are earning $\leq\$50K$. There are 14 attributes: 8 categorical and 6 continuous attributes. There are two attributes corresponding to the race and sex of an individual that can be used to define the protected groups.

\begin{table}
\parbox{.5\linewidth}{
\centering
\caption{Fairness Report}
\label{tab:1}
\begin{tabular}{llllll}
\toprule
&  DPD &  DPR &  EoD &  EoP \\

\midrule
Real Data  &  0.18 &  0.80 &  0.11 &  0.09 \\
Balanced Data &  0.29 & 0.64 & 0.20 & 0.20 \\
CTGAN & 0.05 & 0.95 & 0.04 & 0.042 \\
GaussianCopula & 0.13 & 0.86 & 0.16 & 0.074 \\
CopulaGAN & 0.01 & 0.99 & 0.02 & 0.01 \\
PATE-GAN & 0.45 & 0.46 & 0.39 & 0.39 \\
\bottomrule
\end{tabular}
}
\hfill
\parbox{.5\linewidth}{
\centering
\caption{How frequently model is predicting an individual earning $\geq \$50K$ for the groups (male and female) in test set.}
\label{tab:2}
\begin{tabular}{llll}
\toprule
 &  Male (7677)  & Female (3639) \\
\midrule
Ground Truth & 2346 (30\%) & 412 (11\%) \\
Real Data & 2125 (27\%) & 347 (9\%)  \\
Balanced Data  & 3702 (48\%) & 699 (19\%) \\
CTGAN & 736 (9.5\%) & 170 (4\%) \\
GaussianCopula & 1325 (17\%) & 151 (4\%)  \\
CopulaGAN & 305 (3.9\%) & 121 (3.3\%)  \\
PATE-GAN & 4790 (62\%) & 646 (17.7\%)  \\
\bottomrule
\end{tabular}
}
\end{table}

\subsection{Bias Induced from the Dataset}

The model's performance in terms of group fairness metrics discussed above is presented in Table \ref{tab:1}. We can see that the values in Table \ref{tab:1} indicate that the model is unfair as DPD and EoD are greater than 0.1. To further investigate the cause of the bias, we visualize the correlation matrix of the dataset's attributes and found a high correlation between some of the allowed variables (non-protected) and gender (protected variables) as illustrated in Figure \ref{balanced_data_corr}. Next, we trained the Random Forest classifier on the balanced dataset by maintaining the same class ratio. However, from Table \ref{tab:1} we can observe that the balanced dataset has no impact as the model remains unfair. From Figure \ref{balanced_data_corr} we can observe non-protected attributes still have a high correlation with protected attributes.

Further, we analyze individual fairness in the model outcome. Here, we evaluate how frequently the model predicts an individual as "earning more than $\$50k$ per annum" irrespective of gender. Table \ref{tab:2} shows the performance in terms of individual fairness. We can observe that 30\% of Male and 11\% Female individuals earn more than $\$50k$ per annum. Hence the dataset has more instances of male earning more than $\$50k$ per annum than females. When we train the model on the Adult dataset after removing the gender (protected attribute) information, the proportion of predicting high-earning individuals for males is higher than for females. The model trained on real dataset predicts 27\% male and 9\% female individuals as earning more than $\$50k$ per annum. However, the model trained on balanced dataset predicts 48\% male and 19\% female individuals as earning more than $\$50k$ per annum. This difference in predicting high-earning individuals between both the groups increases even after training the model on the class balanced dataset.

\subsection{Impact of Synthetic data on Bias and Fairness}
We further study the impact of synthetic data on the model's fairness. To generate the synthetic data for our experiments, we have used Synthetic Data Vault (SDV) library\footnote{https://sdv.dev/}. SDV is an open-source library to help end-users easily generate synthetic Data for different data modalities, including single table, multi-table, and time-series data. The SDV Python package includes a comprehensive set of generative models and evaluation framework that facilitates the task of generating evaluating the quality of Synthetic Data. From the SDV library, we have used CTGAN, GaussianCopula~\citep{masarotto2012gaussian}, and Copula model for data generation. We also explored the effect of differentially private algorithms for data generation. We implemented the PATE-GAN framework from the paper~\citep{jordon2018pate} with $\epsilon=2$ privacy budget, where smaller $\epsilon$ implies greater privacy.

\begin{figure*}
\centering
\begin{tabular}{c c c}
\begin{subfigure}{0.3\textwidth}
    \centering
    \includegraphics[width=37mm,scale=0.5]{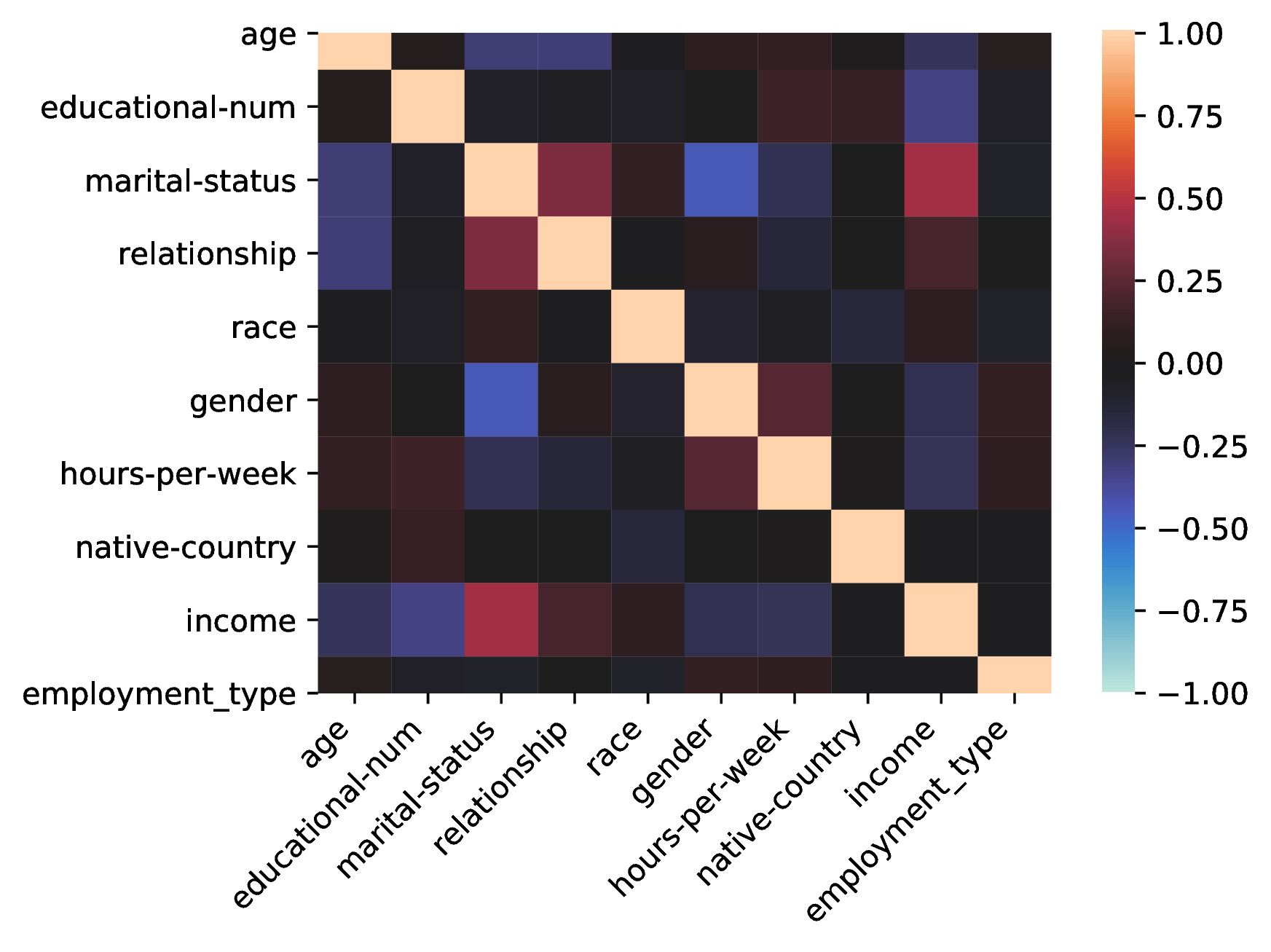}
    \caption{Real Data}
    \label{real_data_corr}
\end{subfigure}
\begin{subfigure}{0.3\textwidth}
    \centering
    \includegraphics[width=37mm,scale=0.5]{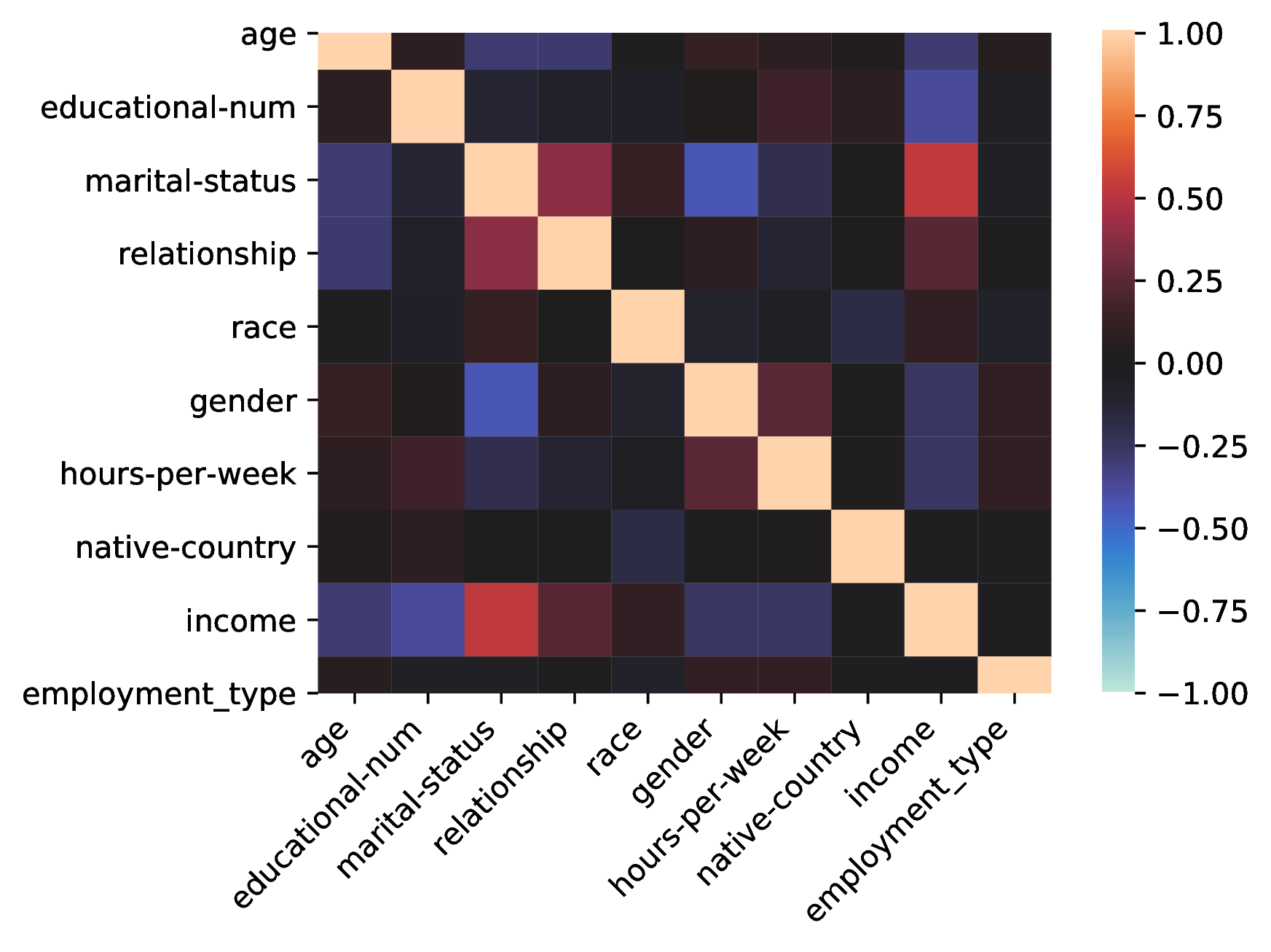}
    \caption{Balanced Data}
    \label{balanced_data_corr}
\end{subfigure} 
\begin{subfigure}{0.3\textwidth}
    \centering
    \includegraphics[width=37mm,scale=0.5]{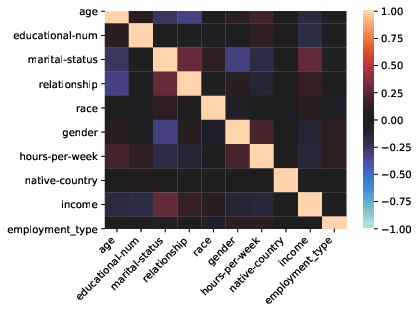}
    \caption{Gaussian Copula}
    \label{gaussian_copula_corr}
\end{subfigure}\\
\begin{subfigure}{0.3\textwidth}
    \centering
    \includegraphics[width=37mm,scale=0.5]{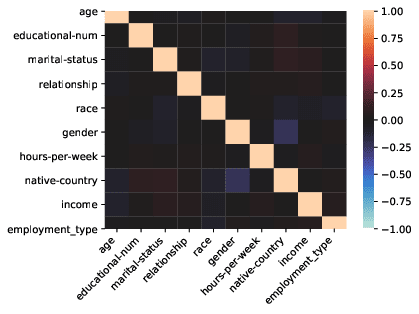}
    \caption{CTGAN}
    \label{ctgan_corr}
\end{subfigure}
\begin{subfigure}{0.3\textwidth}
    \centering
    \includegraphics[width=37mm,scale=0.5]{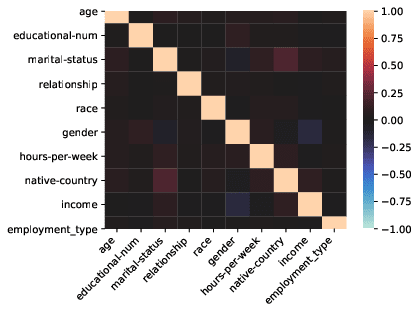}
    \caption{CopulaGAN}
    \label{copula_gan_corr}
\end{subfigure}
\begin{subfigure}{0.3\textwidth}
    \centering
    \includegraphics[width=37mm,scale=0.5]{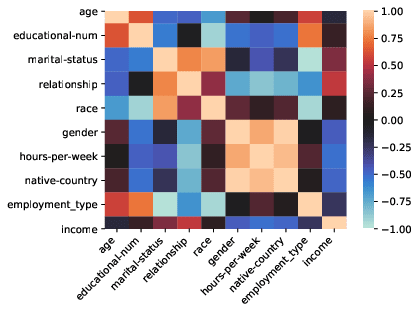}
    \caption{Pate-GAN}
    \label{pate_gan_corr}
\end{subfigure}
\end{tabular}
\caption{Represents the heatmap of the correlation matrix in synthetic data obtained from different generative models. Each heatmap of the correlation matrix shows a correlation coefficient between the variables. The higher the color variation in a cell, the higher is the relationship between the variables.}
\label{correlation_matrix}
\end{figure*}

Table \ref{tab:1} provides details of performance on synthetic data in terms of fairness metrics. The synthetic data generated by methods like GaussianCopula, CTGAN, and CopulaGAN, significantly reduces the model's disparity. The improvement in DPD and EoP indicates the model correctly classifies the same proportion of positive outcomes in both groups. However, a differentially private version of synthetic data (PATE-GAN) amplifies bias and unfairness in the model. This behavior can be explained by the correlation of the attributes in the synthetic data shown in Figure \ref{correlation_matrix}. In the synthetic data generated from CTGAN and CopulaGAN, all the attributes are weakly correlated and loosely dependent upon protected attributes (gender). In PATE-GAN, the attributes are highly correlated.  

Next, we evaluate the model performance in terms of individual fairness as shown in Table \ref{tab:2}. We find that training the model on synthetic data  from CTGAN and CopulaGAN, the proportion of predicting high-earning individuals is almost the same for both groups. In contrast, PATE-GAN model predicts the higher-earning individuals more frequently for the male group. The disparities in outcomes for PATE-GAN is due to random noise added in their differentially private mechanism. Based on all the observations it is evident that the model trained on weakly correlated data is fairer and less biased. Hence an effective way to improve the fairness of the machine learning systems by introducing the use of less correlated synthetic data.

\section{Conclusion}
With the widespread adoption of synthetic data for being compliant and preserving privacy, the study evaluates its impact on bias amplification. Specifically, we studied disparity in machine learning model output when trained using synthetic data. Further, a differentially private version of GANs known as PATE-GAN is also studied in quantifying the impact of bias amplification. Experimental evaluation reveals that differentially private versions of the synthetic dataset possess more disparity as evident from fairness metrics. This effect is attributed to high cross-correlation in generated features. The outcome of the study will ensure that synthetic data is carried in responsible ways while accounting for the model disparity. We encourage future research directions to evaluate the impact of the privacy budget on model disparity. This further opens up the avenue to explore the trisection of utility, privacy, and fairness in synthetic datasets.

\clearpage
\bibliography{iclr2021_conference}
\bibliographystyle{iclr2021_conference}

\clearpage
\appendix
\section*{Supplemental}

\setcounter{section}{0}

\section{Dataset Preprocessing}
We assess the model's bias and fairness on the Adult dataset. Before training the model, we did the numerical encoding of the categorical variables. We have removed the samples with missing values and weakly correlated attributes  "capital-gain," "capital-loss," "fnlwgt" from the dataset. We have merged the marital-status attribute values "Divorced, Married-spouse-absent, Never-married, and Separated" into "Single" and "Married-AF-spouse, Married-civ-spouse" into "Couple." Finally, to ensure that the data do not influence the model's unfairness based on gender, we removed the gender attribute from the training.

\section{Fairness Metrices}
Below are the evaluation metrics to measure group fairness:\\
(i) Demographic Parity Difference (DPD): DPD is the absolute difference between the true positive rate of each group. DPD defines the model's fairness as the likelihood of a positive outcome should be the same for each group~\citep{verma2018fairness}. (ii) Demographic Parity Ratio (DPR): DPR is the ratio of positive outcomes. (iii) Equality of Odds (EoD): EoD suggests that the true positive rate and the false positive rate should be the same for each group. EoD takes the absolute difference between the true positive rate of different groups or the absolute difference between the false positive rate of different groups, whichever is maximum. (iv) Equality of Opportunity (EoP): EoP focuses only on the true positive rate of the model~\citep{hardt2016equality}. It is the difference of true positive rates among the groups. \\
Some cut-off values have been defined in the literature for these metrics, suggesting whether models are fair or not. For DPD, EoD, and EoP, if the absolute value is smaller than 0.1, the model can be considered fair, whereas for DPR fairness range for this metric is between 0.8 and 1.25.

\section{Fairness Analysis}
\begin{figure*}
\centering
\begin{tabular}{c c}
\begin{subfigure}{0.5\columnwidth}
    \centering
    \includegraphics[width=70mm,scale=1]{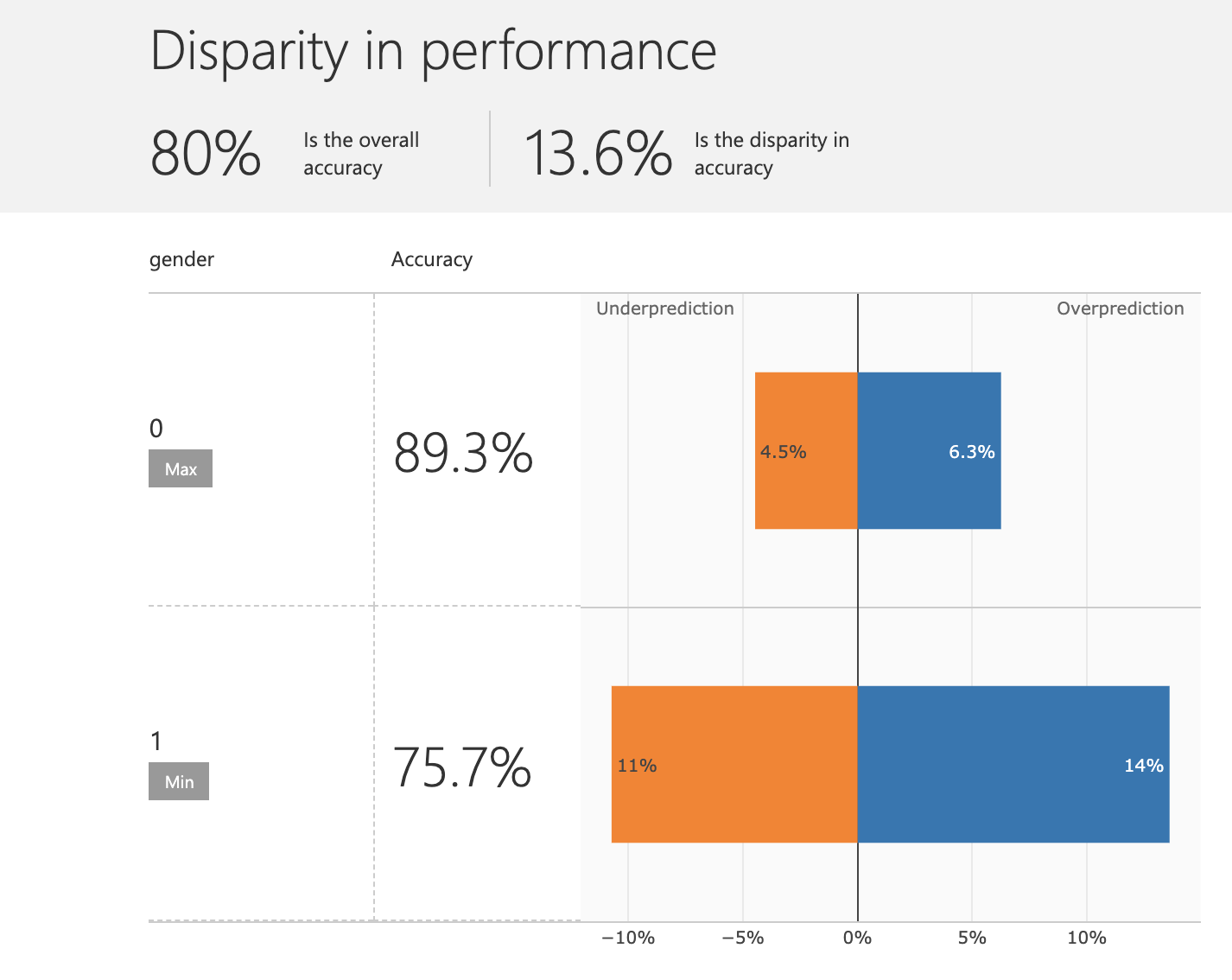}
    \caption{Real Data}
    \label{real2}
\end{subfigure}
\begin{subfigure}{0.5\columnwidth}
    \centering
    \includegraphics[width=70mm,scale=1]{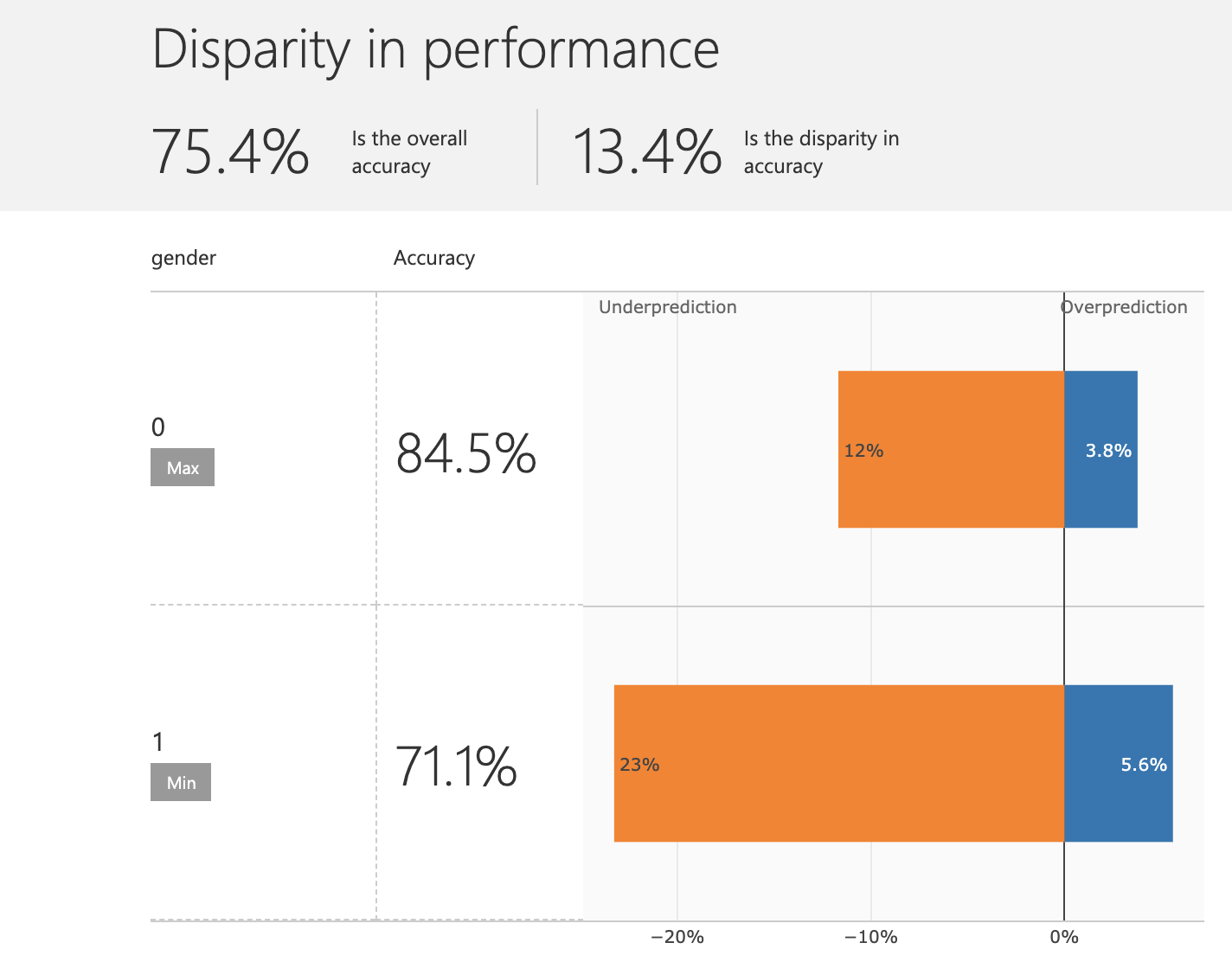}
    \caption{Balanced Data}
    \label{balanced2}
\end{subfigure} \\
\begin{subfigure}{0.5\columnwidth}
    \centering
    \includegraphics[width=70mm,scale=1]{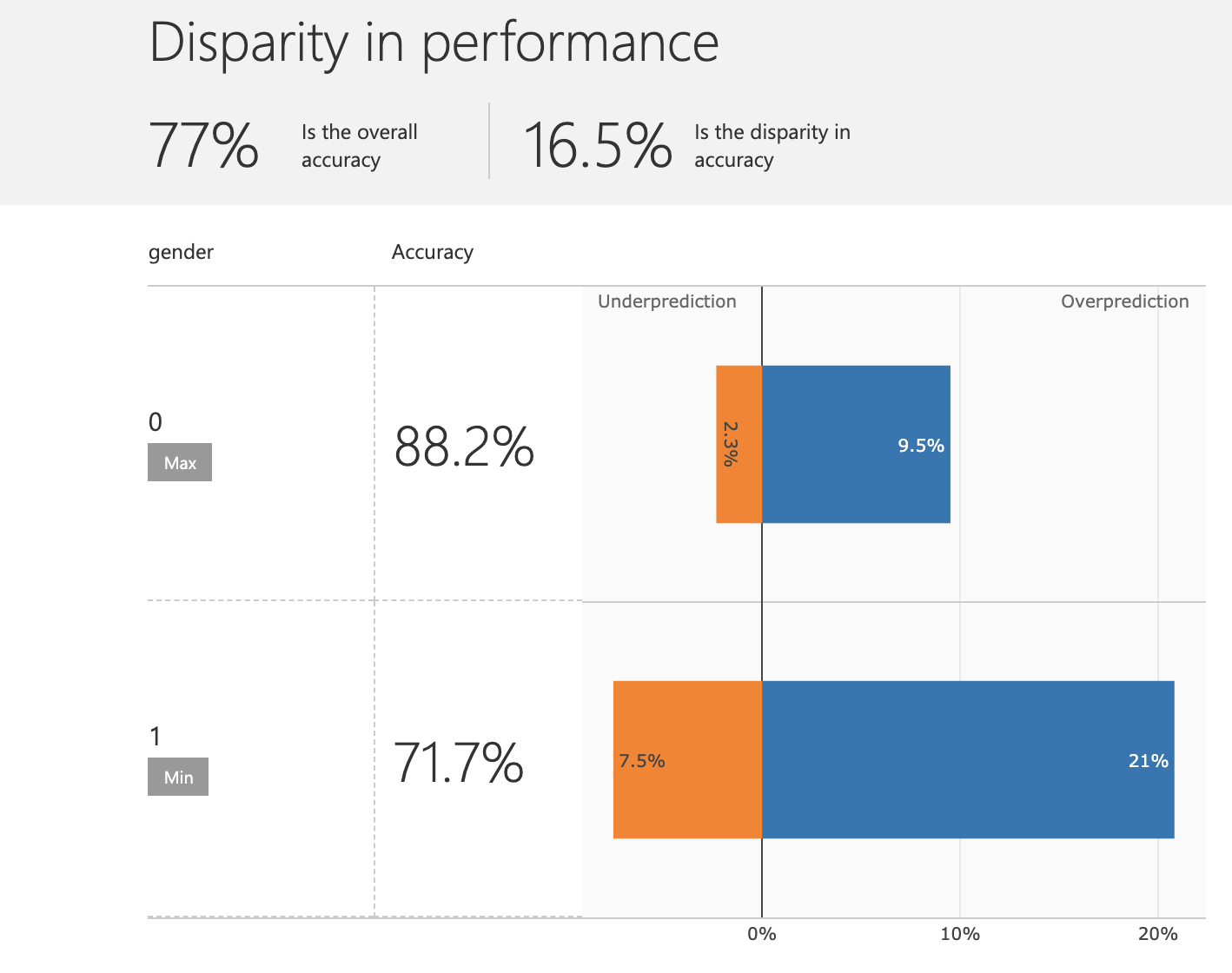}
    \caption{Gaussian Copula}
    \label{gaussian_copula2}
\end{subfigure}
\begin{subfigure}{0.5\columnwidth}
    \centering
\includegraphics[width=70mm,scale=1]{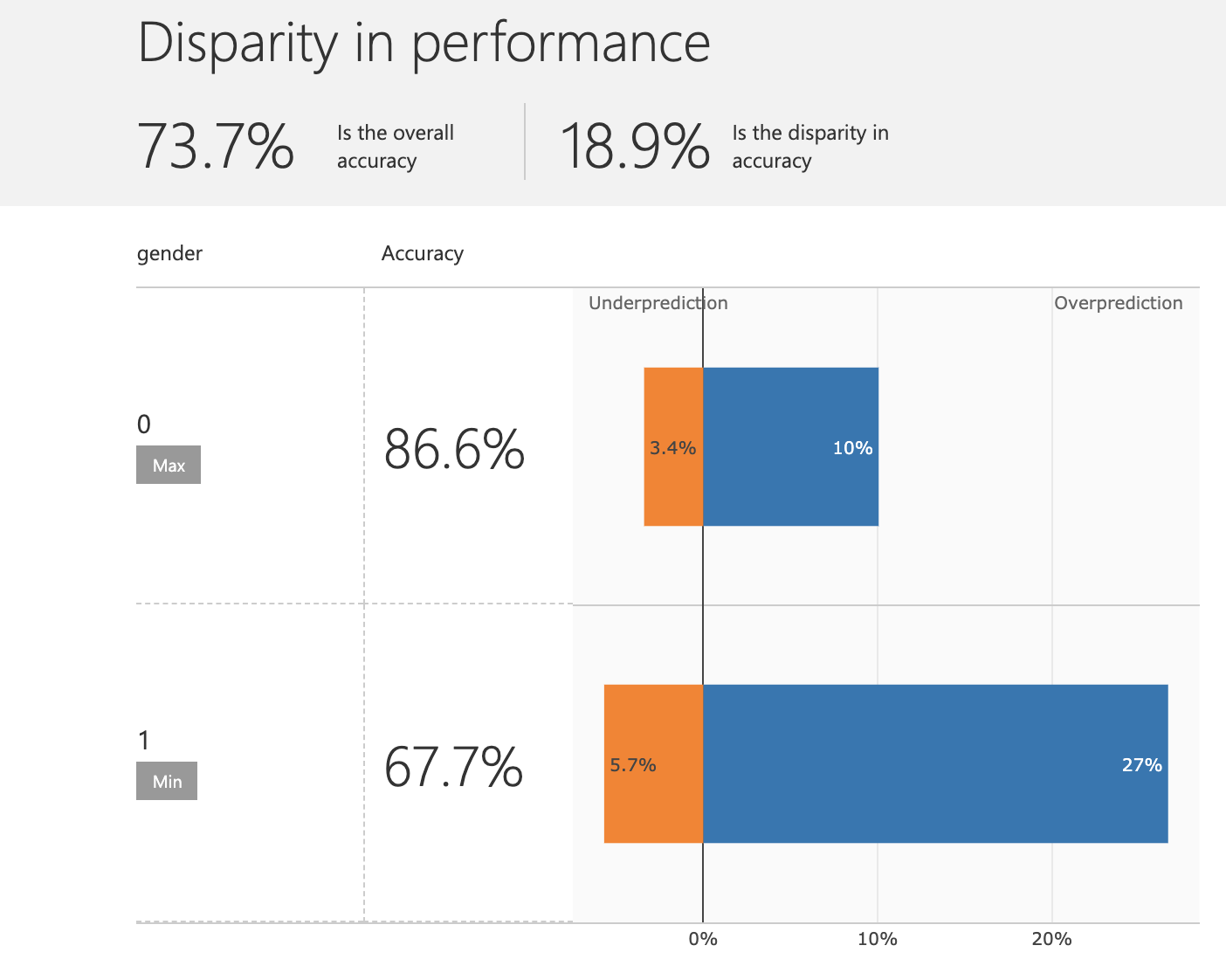}
    \caption{CTGAN}
    \label{ctgan2}
\end{subfigure} \\
\begin{subfigure}{0.5\columnwidth}
    \centering
    \includegraphics[width=70mm,scale=1]{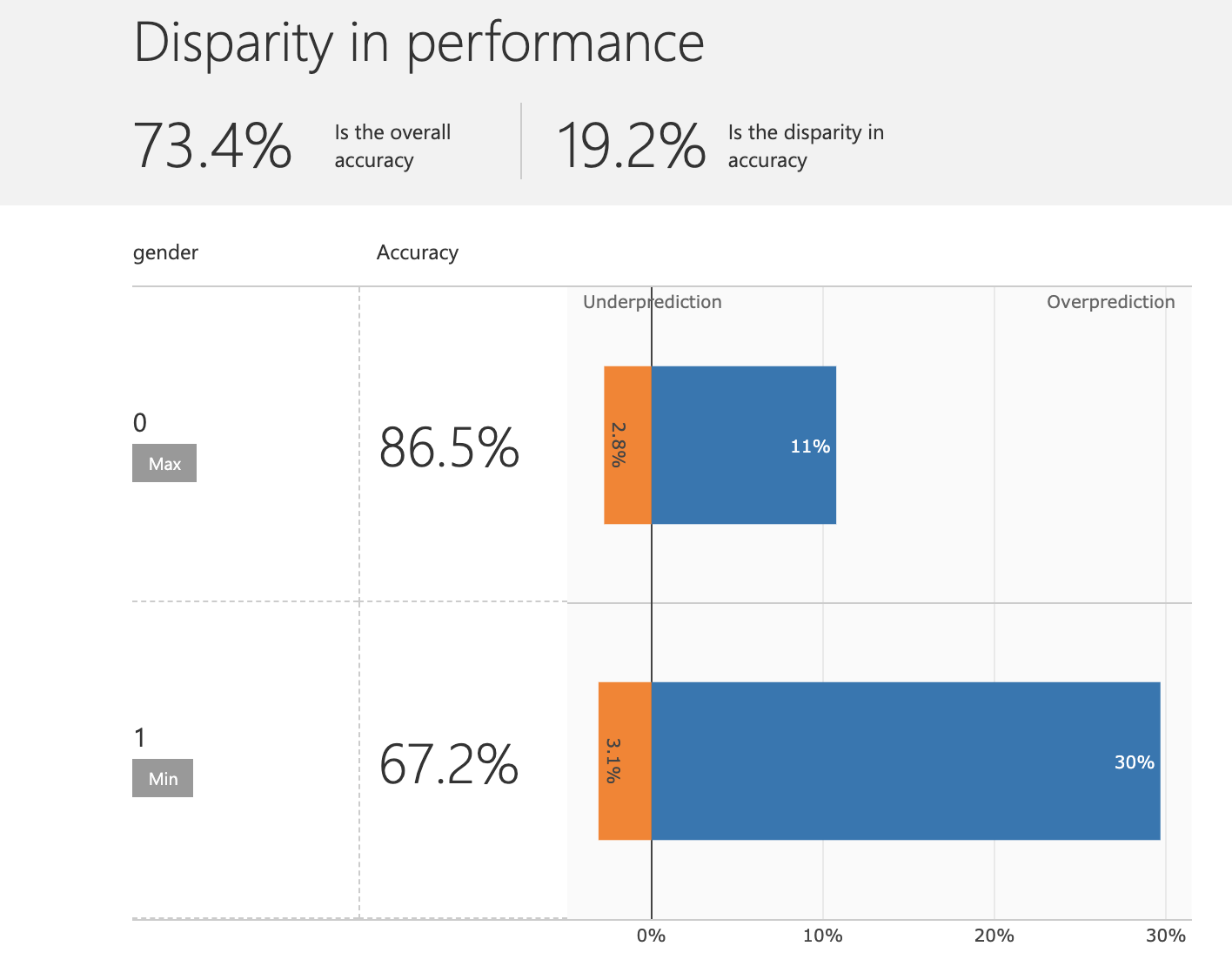}
    \caption{CopulaGAN}
    \label{copula_gan2}
\end{subfigure}
\begin{subfigure}{0.5\columnwidth}
    \centering
    \includegraphics[width=70mm,scale=1]{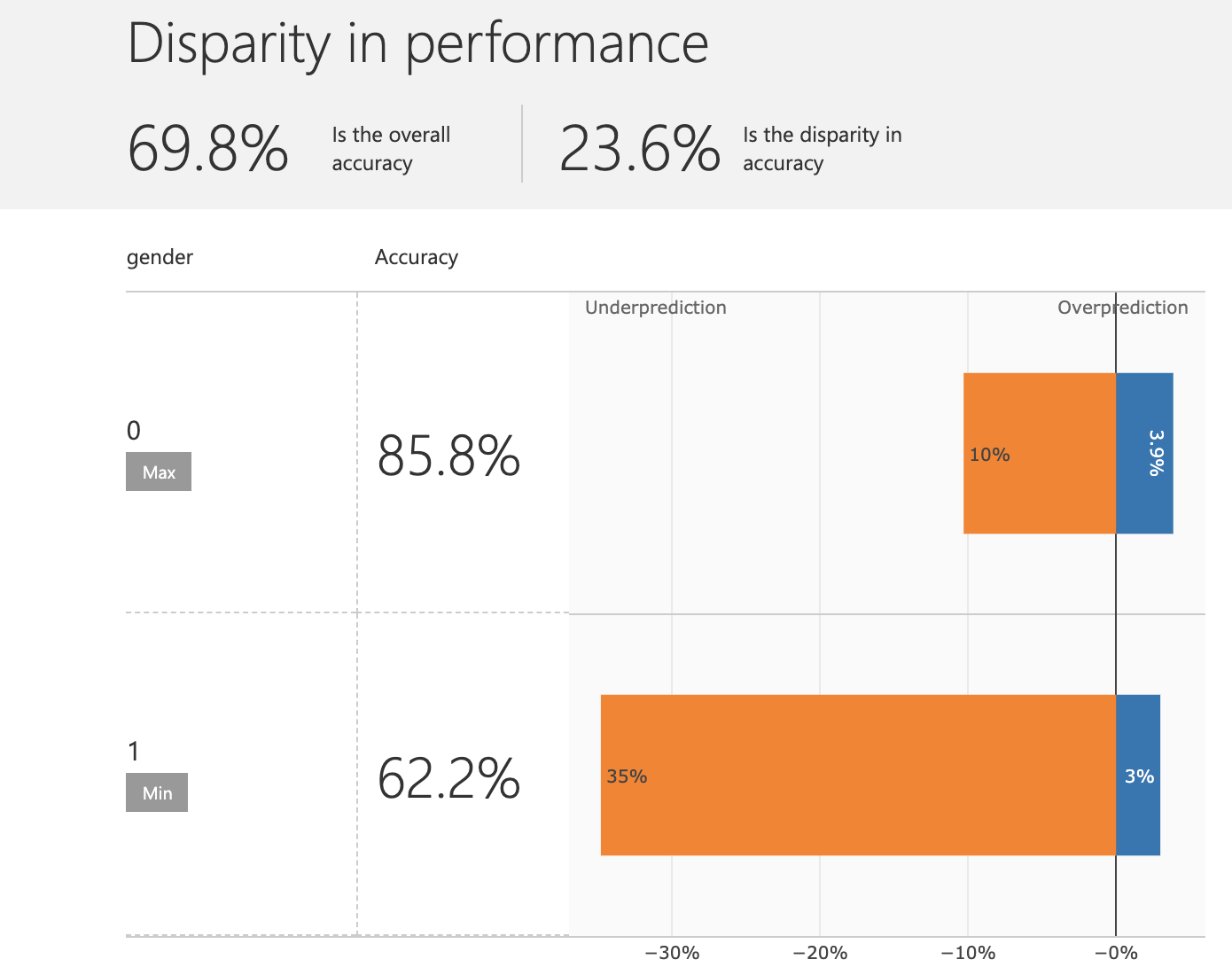}
    \caption{Pate-GAN}
    \label{pate_gan2}
\end{subfigure}

\end{tabular}
\caption{Demonstrate the difference in performance of the models for the groups (Female - Top and Male - Bottom). The bar chart shows distribution of errors in each groups. Errors are split into overprediction error (predicting 1 when the true label is 0), and underprediction error (predicting 0 when the true label is 1).}
\label{performance_result}
\end{figure*}

Additionally, we evaluate the model fairness based on classification accuracy as shown in Figure~\ref{performance_result}. We have trained the model on the real Adult dataset using Random Forest. There is a deviation in the overall accuracy and group accuracy as illustrated in Figure \ref{real2}. The classifier's overall accuracy is 80\%, while the accuracy for females and males are 89.3\% and 75.7\%, respectively. There is a significant difference in performance between the two groups, indicating the unfairness in our model. We have not used the gender attribute to train the classifier; still, there is a significant difference in performance across the two gender groups. This is primarily because some non-protected variables may possess a high correlation with gender (protected variable). Empirically, we find that maintaining the class ratio in the dataset does not solve the model's disparity, as shown in Figure\ref{balanced2}.

Similar results can be observed when we trained on synthetic data, where they exhibit deviation in the overall accuracy between the groups implying unfairness. Earlier, we have observed that in terms of fairness metrics, synthetic data has shown an improvement. This is primarily because the accuracy metric focuses on both the positive and negative outcomes of the model's prediction. In contrast, the fairness metric only focuses on the positive outcomes of the model's prediction. Also, from the Figure \ref{ctgan2} and \ref{copula_gan2}, we can see that both the CTGAN and CopulaGAN miss-classify the same proportion of positive class among the group, i.e., the under-prediction error (model predicting 0 when the true label is 1) is similar for both the groups. Whereas in PATE-GAN, the male group's misclassification rate is much higher than the female group, as shown in Figure \ref{pate_gan2}. This result supports our previous finding in the paper.

\end{document}